\documentclass{article}

\usepackage[final]{funasr_tech_report}

\usepackage[utf8]{inputenc} % allow utf-8 input
\usepackage[T1]{fontenc}    % use 8-bit T1 fonts
\usepackage{hyperref}       % hyperlinks
\usepackage{url}            % simple URL typesetting
\usepackage{booktabs}       % professional-quality tables
\usepackage{amsfonts}       % blackboard math symbols
\usepackage{authblk} % for \affil command
\usepackage{adjustbox}      % for adjustbox environment
\usepackage{nicefrac}       % compact symbols for 1/2, etc.
\usepackage{microtype}      % microtypography
\usepackage{xcolor}         % colors
\usepackage{graphicx}
\usepackage{subcaption} % 用于子图支持
\usepackage{amsfonts}       % blackboard math symbols
\usepackage{amsmath}        % math environments and more
\usepackage{cite}
\usepackage{multicol}
\usepackage{multirow}
\usepackage{enumitem}
\usepackage{pifont}
\newcommand{\cmark}{\ding{51}}%
\newcommand{\xmark}{\ding{55}}%

\title{Fun-ASR Technical Report}

\author{Tongyi Fun Team}
\affil{Alibaba Group}
\affil{Fun-ASR@list.alibaba-inc.com}
\begin{document}

\maketitle

\begin{abstract}
In recent years, automatic speech recognition (ASR) has witnessed transformative advancements driven by three complementary paradigms: \textit{data scaling}, \textit{model size scaling}, and \textit{deep integration with large language models (LLMs)}. However, LLMs are prone to hallucination, which can significantly degrade user experience in real-world ASR applications.
In this paper, we present \textbf{Fun-ASR}, a large-scale, LLM-based ASR system that synergistically combines massive data, large model capacity, LLM integration, and reinforcement learning to achieve state-of-the-art performance across diverse and complex speech recognition scenarios. Moreover, Fun-ASR is specifically optimized for practical deployment, with enhancements in streaming capability, noise robustness, code-switching, hotword customization, and satisfying other real-world application requirements. Experimental results show that while most LLM-based ASR systems achieve strong performance on open-source benchmarks, they often underperform on real industry evaluation sets. Thanks to production-oriented optimizations, Fun-ASR achieves SOTA performance on real application datasets, demonstrating its effectiveness and robustness in practical settings. The code and models are accessible at \url{https://github.com/FunAudioLLM/Fun-ASR}
\end{abstract}

\begin{figure}[h!]
    \centering
    \includegraphics[width=0.75\linewidth]{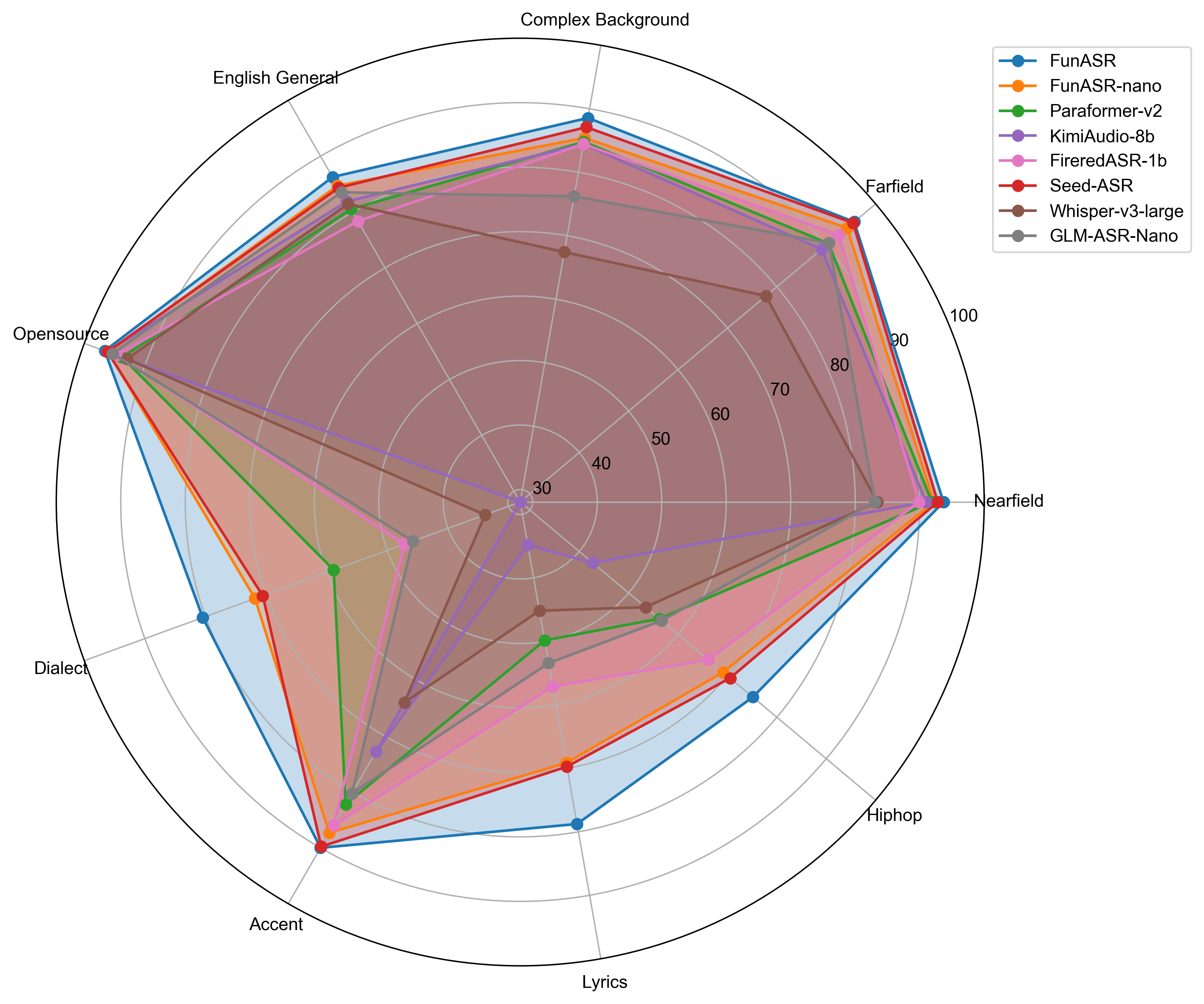}
    \caption{Performance comparison (in Accuracy) between our Fun-ASR (7.7B) and Fun-ASR-nano (0.8B) and top-tier ASR and speech-text multimodal models.}
    %, including Paraformer-v2, Kimi-Audio (8B), FireRedASR (1.1B), and Seed-ASR ($\sim$12B).}
    \label{fig:Performance}
\end{figure}

\section{Introduction}
In recent years, under the paradigms of data scaling, model size scaling, and deep integration with large language models (LLMs), automatic speech recognition (ASR) has undergone transformative advancements in both methodology and real-world application scenarios. These three complementary paradigms have collectively driven the evolution of ASR systems from traditional neural-network-based approaches to modern large-model-based architectures, culminating in state-of-the-art (SOTA) performance across diverse acoustic and linguistic conditions.

Data scaling has proven to be a fundamental driver of ASR improvements. The seminal work on Whisper~\citep{DBLP:conf/icml/RadfordKXBMS23} provides empirical evidence that ASR performance exhibits a strong positive correlation with the scale of the training data. Their comprehensive experiments demonstrated that increasing the training data volume from 3K hours to over 680K hours results in more than 20-point drop in the English word error rate (WER). This significant improvement underscores the critical role of data diversity and quantity in ASR development, as larger datasets enable models to capture a more comprehensive representation of linguistic and acoustic variations across different languages, accents, speaking styles, and environmental conditions. The availability of massive, multilingual speech corpora has become a cornerstone of modern ASR development, with the most successful systems now leveraging datasets spanning tens of millions of hours.

Model size scaling, particularly the increase in the number of model parameters, has further amplified the benefits of data scaling. The scaling laws observed in large language models (LLMs) have been extended to speech recognition, where increasing model size while maintaining data scaling has yielded substantial performance gains. For example, the Whisper model family demonstrated that increasing the model size from 38M to over 1500M led to more than 40-point WER reduction in multilingual ASR. This synergy between data and model scaling has been pivotal in the development of modern ASR systems, with the largest Whisper variants achieving WERs on par with human transcriptions. 

The third paradigm, deep integration with LLMs, represents a paradigm shift in the ASR methodology. Rather than treating ASR as a standalone task, this approach leverages the rich linguistic knowledge and contextual understanding of LLMs to enhance speech recognition. Models such as Seed-ASR~\citep{DBLP:journals/corr/abs-2407-04675} and FireRedASR~\citep{DBLP:journals/corr/abs-2501-14350} have demonstrated that incorporating LLMs can significantly improve ASR performance, particularly in resolving semantic ambiguities and generating more coherent and contextually appropriate transcriptions. These models effectively bridge the gap between speech and text understanding. 

Building upon these significant advancements, we propose \textbf{Fun-ASR}, a large-scale LLM-based ASR system trained on large-scale data. Fun-ASR exhibits the following key characteristics:

\begin{itemize}[leftmargin=*,noitemsep]
%%%%\begin{itemize}
    \item \textbf{Scaling and Innovative LLM Integration.} Fun-ASR is designed to harness the synergistic benefits of data scaling, model size scaling, and LLM integration. %\wen{Please concisely summarize the status of our data scaling and model size scaling, that is, how much training data we used and our model size. Also, please concisely summarize our \textbf{innovations} in the model architecture, that is, the innovative integration with LLMs.}
    \item {\bf State-of-the-art speech recognition accuracy.} Through synergistic advancements in data scaling, model size scaling, and innovative architectural integration with LLMs, Fun-ASR achieves unprecedented recognition accuracy across diverse linguistic and acoustic domains, establishing a new state of the art for ASR systems. Our comprehensive evaluations demonstrate that Fun-ASR substantially outperforms both our previous small-scale models and leading ASR systems in industry, in terms of critical metrics across multiple challenging benchmarks. %\wen{Please add representative quantitative results here to support this statement.}
    % 补充有代表性的结果，特别是相比其他产品有明显优势的结果
    %@ zhendong
    \item {\bf Optimization for practical production usage.}
Beyond achieving state-of-the-art performance on standardized benchmarks, Fun-ASR is meticulously engineered to meet the complex demands of real-world deployment scenarios, with a particular focus on practical usability, reliability, and user experience. We implement a comprehensive suite of optimizations in multiple dimensions, each addressing specific challenges encountered in commercial applications. 
(1)
First, we implement \textbf{a highly efficient streaming ASR architecture} for Fun-ASR that supports real-time processing with minimal latency, enabling seamless integration into live applications such as video conferencing, live captioning, and voice-controlled devices. 
(2)
Second, we enhance \textbf{noise robustness} substantially through a multi-stage approach. 
(3)
Third, we implement \textbf{advanced code-switching capabilities} that seamlessly handle transitions between Chinese and English within the same utterance, which is critical for multilingual users in global business environments.
(4)
Fourth, we integrate \textbf{customizable hotword recognition} that allows users to define domain-specific terms or phrases for enhanced recognition accuracy. This feature is particularly valuable in specialized domains such as healthcare, enterprise, and automotive technology. %%%%\wen{Please add quantitative results for customizable hotword recognition here.} %补充代表性结果

To thoroughly evaluate these production-oriented optimizations, we develop a \textbf{comprehensive evaluation protocol} that includes both standardized benchmarks and real-world usage scenarios. This protocol encompasses %补充测试集情况
distinct test sets, each simulating different application contexts. %补充测试集覆盖的领域
Our evaluation results reveal that Fun-ASR not only excels in recognition accuracy but also provides superior practical performance. These results demonstrate that Fun-ASR successfully bridges the gap between academic research and commercial production readiness, offering a comprehensive solution to address real-world speech recognition challenges.
\end{itemize}

This report is organized as follows. Section~\ref{sec:architecture} and Section~\ref{sec:data} introduce the model architecture and the training data. Section~\ref{sec:training} elaborates the training paradigm. Section~\ref{sec:production-optimization} describes how we implement critical production-oriented optimizations.
%%%%for different application scenarios. 
Experiments are presented in Section~\ref{sec:evaluation}, followed by discussions on the limitations of this work and our future plans.

\section{Model Architecture}
\label{sec:architecture}
% @ zhifu

\begin{figure}[h]
    \centering
    \includegraphics[width=0.9\linewidth]{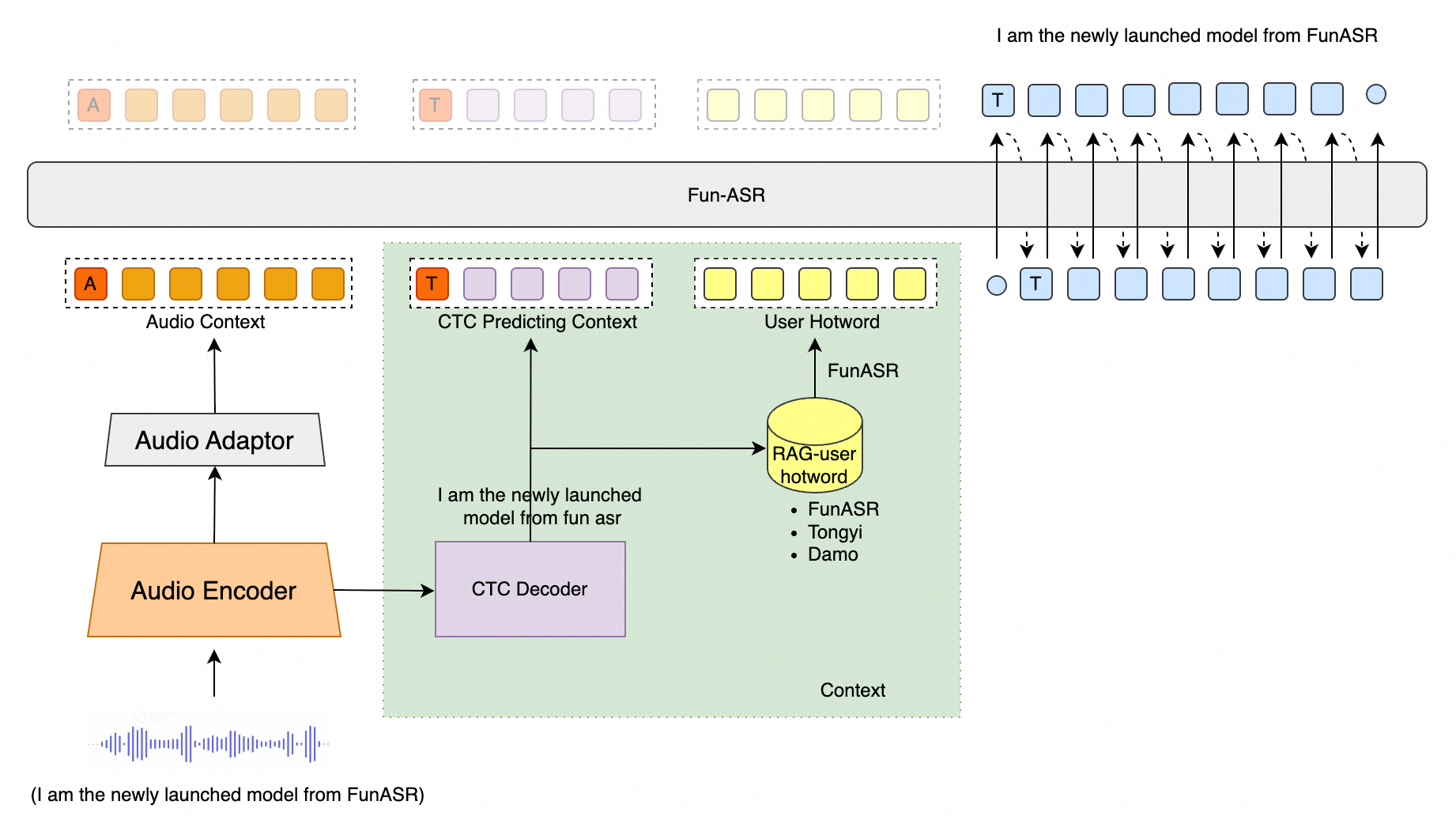}
    \caption{Overview of the Fun-ASR model architecture.}
    \label{fig:overview}
\end{figure}

Fun-ASR comprises four key components: (1) an audio encoder that extracts representations from the input speech, implemented as multiple layers of transformer encoder, (2) an audio adaptor that connects the audio encoder output with the LLM, implemented as two layers of transformer encoder, (3) a CTC decoder that is built upon the audio encoder to obtain the initial recognition hypothesis, which will be used for hotword customization, as described in Section~\ref{subsec:hotword}, and (4) an LLM-based decoder that produces output based on the audio condition and the CTC prediction. 

To address varying constraints on computational resources and requirements on inference efficiency, we propose two models with different model sizes: \textbf{Fun-ASR} and \textbf{Fun-ASR-nano}. Fun-ASR comprises an audio encoder with 0.7B parameters and an LLM-based decoder with 7B parameters, aiming for high recognition accuracy; Fun-ASR-nano comprises an audio encoder with 0.2B parameters and an LLM-based decoder with 0.6B parameters, seeking to strike a balance between accuracy and efficiency to meet the demands of low-computational-resource scenarios.

\section{Data}
\label{sec:data}

\subsection{Pre-taining Data}
\label{subsec:pretraining-data}
The pre-training dataset comprises approximately \textbf{tens of millions hours} of audio data, including both unlabeled audio and labeled audio-text data. The unlabeled audio data span a broad range of real-world scenarios in domains such as artificial intelligence, biotechnology, e-commerce, education, entertainment, finance, and mobility. For labeled data, a comprehensive data processing pipeline is employed, which incorporates voice activity detection (VAD), pseudo-label generation by multiple ASR systems (such as Paraformer-V2~\citep{DBLP:journals/corr/abs-2409-17746}, Whisper,  and SenseVoice~\citep{DBLP:journals/corr/abs-2407-04051}), followed by inverse text normalization (ITN). The primary languages in the labeled dataset are Chinese and English.

\subsection{Supervised Fine-tuning Data}
\label{subsec:sft-data}
The supervised fine-tuning (SFT) data consist of approximately \textbf{millions of hours} of data, including: human-transcribed data, pseudo-labeled data, environmental noise data, TTS generated data using CosyVoice3~\citep{DBLP:journals/corr/abs-2505-17589}, simulated streaming data, noise augmented data, and hotword customized data.

\section{Training}
\label{sec:training}

\subsection{Pre-training of Audio Encoder}
\label{subsec:pre-training}
\begin{figure}[h]
    \centering
    \includegraphics[width=0.9\linewidth]{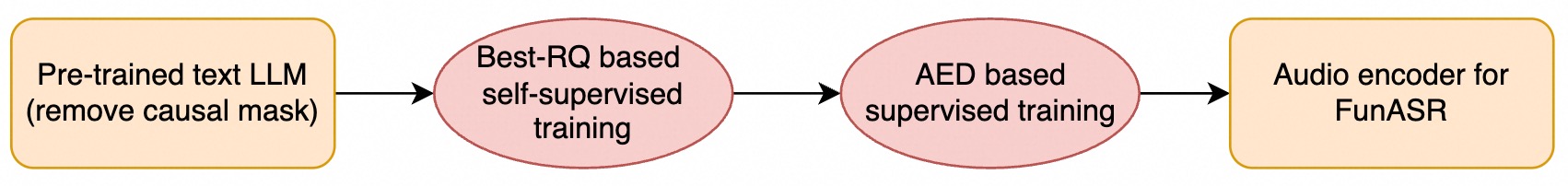}
    \caption{The pre-training pipeline for the audio encoder.}
    \label{fig:encoder}
    % \vspace{-8mm}
\end{figure}

To develop a robust and effective audio encoder for integration into an LLM-based ASR (LLM-ASR) system, we adopt two complementary approaches, as shown in Figure~\ref{fig:encoder}. These strategies aim to leverage both self-supervised and supervised learning paradigms to produce high-quality speech representations that can be effectively aligned with the linguistic knowledge in the LLM.

{\bf Stage 1: Self-supervised pre-training via Best-RQ framework with pre-trained text LLM initialization.}
The first stage leverages self-supervised pre-training through the Best-RQ (BERT-based Speech pre-Training with Random-projection Quantizer) framework~\citep{DBLP:conf/icml/ChiuQZYW22}, a state-of-the-art self-supervised learning method for speech representation learning. Best-RQ operates by masking and reconstructing speech units, using a quantization module to discretize continuous representations. This approach enables the model to learn general-purpose speech representations without requiring any labeled data, making it highly scalable to vast amounts of unlabeled audio data.
Notably, \textbf{a key innovation in our implementation of Best-RQ lies in the initialization strategy}. Drawing inspiration from our recent findings~\citep{DBLP:journals/corr/abs-2409-17750}, where we demonstrate that layers in pre-trained text LLM can effectively initialize the encoder in ASR systems, we initialize the Best-RQ encoder with weights from layers of a pre-trained text LLM—specifically, Qwen3 model~\citep{DBLP:journals/corr/abs-2505-09388}. This cross-modal initialization strategy is based on the hypothesis that the deep linguistic and semantic knowledge encoded in the LLM can provide a beneficial inductive bias for learning speech representations. We observe that initializing Best-RQ with a pre-trained text LLM significantly accelerates the training convergence and improves the quality of the learned speech representations compared to random initializations. 

{\bf Stage 2: Supervised pre-training via attention-based encoder-decoder (AED) framework.}
The second stage involves supervised pre-training of the audio encoder within a conventional attention-based encoder-decoder architecture. This method follows a well-established and empirically validated training paradigm used in our prior works, specifically the SenseVoice-Large model~\citep{DBLP:journals/corr/abs-2407-04051}. In this setup, the encoder is trained end-to-end on large-scale labeled ASR datasets, using standard sequence-to-sequence learning objectives.
The primary objective of this supervised pre-training phase is to obtain an encoder that learns rich acoustic and linguistic features from transcribed speech data. Once trained, the encoder of the AED framework is used to initialize the audio encoder in the downstream LLM-ASR system. This initialization provides a strong starting point for subsequent joint training of the audio and language components, reducing the need for extensive low-level feature learning from scratch and thereby accelerating the training convergence.

\subsection{Supervised Fine-tuning}
\label{subsec:SFT}
% @ keyu changfeng yabin

Supervised Fine-tuning (SFT) comprises four sequential stages:

{\bf Stage 1}: The parameters of the pre-trained audio encoder and the LLM are kept frozen, while the adaptor module is trained to align the audio encoder's output representations with the LLM's semantic space. The training data for this stage is about 200k hours and this stage takes about 70K training steps. 
%\wen{which categories among the SFT data categories are used in this stage?} and this stage takes about 70K training steps. 
% After stage 1, the accuracy of the token prediction is nearly 90\%.

{\bf Stage 2}: The LLM parameters are still kept frozen, while the audio encoder and the adaptor module are trained to learn better semantic representations. This stage uses about 10M hours of low-cost ASR training data and trains one epoch. 

{\bf Stage 3}: The encoder and the adaptor module are frozen, while we update the LLM parameters with Low-Rank Adaptation (LoRA). The purpose of LoRA-based LLM adaptation is to preserve the model's text generation capabilities while ameliorating catastrophic forgetting of the pre-trained knowledge. This LoRA fine-tuning stage uses 20K hours of ASR data.
%\wen{which ASR data among the SFT data?} with 20K training steps.

{\bf Stage 4}: Full-parameter fine-tuning is applied to both the audio encoder and the adaptor, while LoRA is employed to fine-tune the LLM simultaneously.  In this stage, we only use the high quality data, which contains about 3M hours of speech. The transcriptions are evaluated by three different ASR models, including Whisper-Large-V3, FireRed-ASR, and SenseVoice. 
%\wen{Note that when describing models or toolkits we used, please provide their URLs in footnote, for reproducibility}. %\wen{So, did we use the 3M hours of data labeled by the three ASR models here?}

{\bf Stage 5}: As depicted in Figure~\ref{fig:overview}, we add a CTC decoder on top of the audio encoder. During this training stage, the audio encoder is frozen and only the CTC decoder is trained. This CTC decoder is used to obtain the initial recognition hypothesis by greedy search. Then, this one-pass result is used for retrieval-augmented generation (RAG) to obtain the context information.

\subsection{Contextual Supervised Fine-tuning}
\label{subsec:contextual-SFT}
As a content prior, context can effectively help the model identify and disambiguate key text content from easily confusable pronunciation in ASR tasks, and improve the accuracy of long-term continuous recognition in complex scenarios. Consequently, after the SFT training (Section~\ref{subsec:SFT}), we further train Fun-ASR on the contextual and long-duration data to enhance its contextual modeling capability. 

The duration of the audio samples can be up to 5 minutes. For the longer samples, we segment the sample and add the transcript of the previous segment in front of the current audio segment as prompts. Since high-quality contextual audio data is severely limited, we construct over 50K hours of SFT data with contextual content through the following steps. 
%\wen{how many training steps are used for contextual SFT?}

\textbf{Step 1: Keyword Extraction:} To generate contextual information related to the current conversation content, we first extract keywords from its transcript using Qwen3-32B~\citep{yang2025qwen3}. Keywords typically include entities, professional terms, and specific time periods. They are words that ASR systems often fail to recognize.

\textbf{Step 2: Relevant Context Synthesis:} We use the Qwen3-32B model to synthesize contextual content. Given the current conversation content and the extracted keywords, we prompt Qwen3-32B to synthesize multiple, diverse contextual content pieces that align with spoken conversation characteristics. 
For the synthesized context, we then use keyword matching to filter out contextual pieces that do not contain the specified keywords. 
If no keywords are extracted from the current conversation in the previous step, the LLM is prompted to synthesize context based solely on the current conversation content. 

\textbf{Step 3: Irrelevant context combination:} To prevent the model from being overly dependent on the context, we randomly sample five irrelevant contextual pieces for each conversation sample from the dataset and mix them with the synthesized relevant context to form the final contextual SFT training data.

\subsection{Reinforcement Learning}

\subsubsection{The RL Framework for Large Audio-Language Models}

\begin{figure}[htbp]
    \centering
    \begin{subfigure}[b]{0.35\linewidth}
        \centering
        \includegraphics[width=\linewidth]{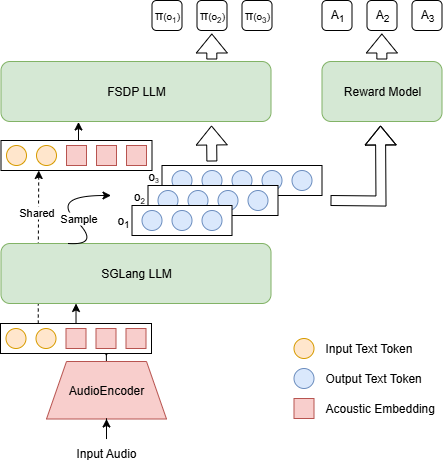} % 替换为你的图片路径
        \caption{The FunRL framework.}
        \label{fig:rl_left}
    \end{subfigure}
    % \hfill
    \hspace{1cm}
    \begin{subfigure}[b]{0.35\linewidth}
        \centering
        \includegraphics[width=\linewidth]{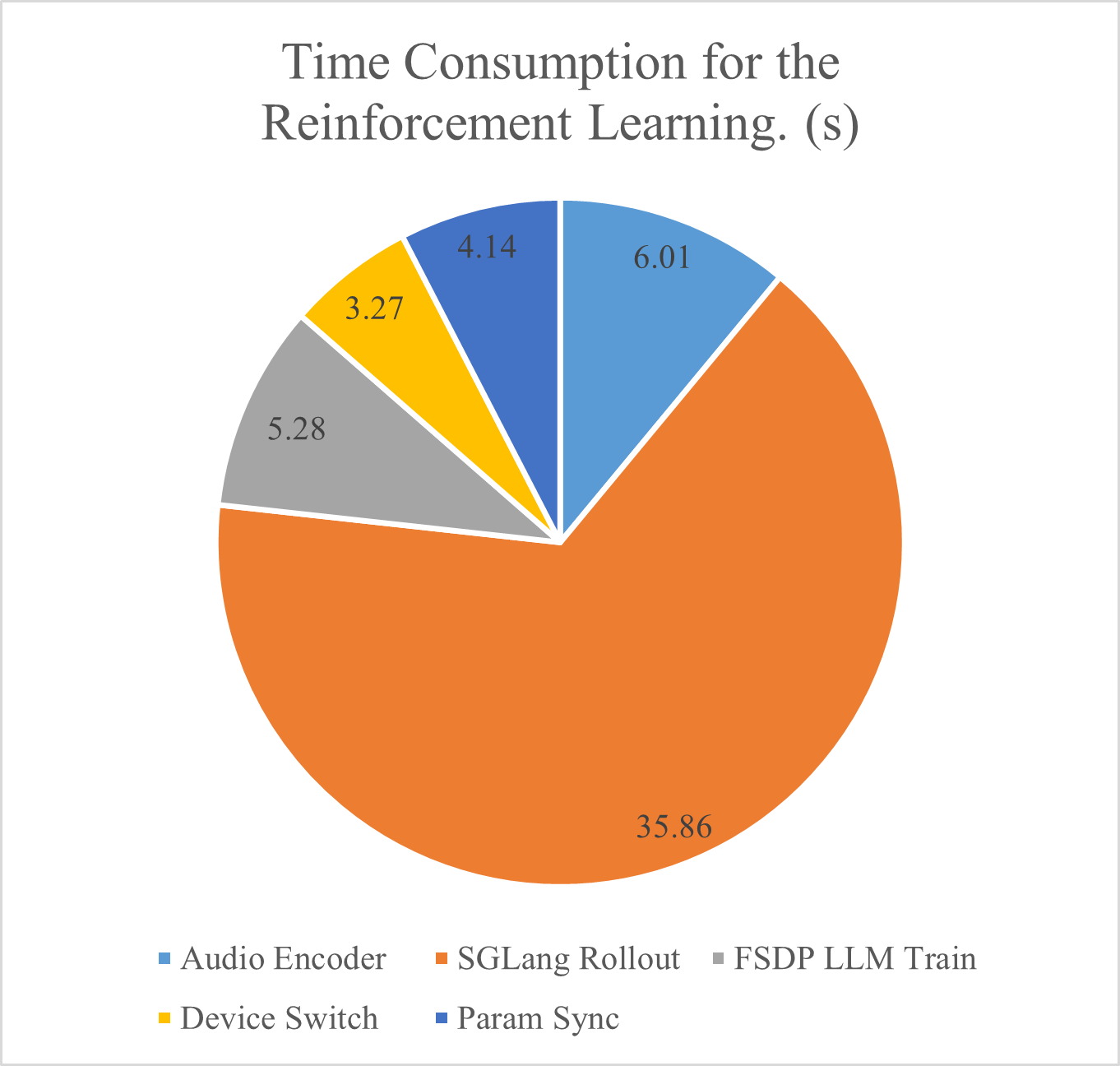} % 替换为你的图片路径
        \caption{Time consumption analysis.}
        \label{fig:rl_right}
    \end{subfigure}
    \caption{The overview and time consumption analysis for our FunRL framework.}
    \label{fig:combined}
\end{figure}

We design \textbf{FunRL}, an efficient reinforcement learning (RL) framework tailored for large audio-language models (LALMs). Different from text LLMs, Fun-ASR, as an LALM, incorporates an audio encoder to convert speech into embeddings—a component that is not natively supported by existing RL frameworks such as Verl~\citep{sheng2024hybridflow} or Trl~\citep{vonwerra2022trl}.
As illustrated in Figure~\ref{fig:rl_left}, FunRL orchestrates the audio encoder, rollout, and policy modules using Ray, enabling them to alternately utilize GPU resources.
In the audio encoder inference stage, all input audio clips are batched and processed through a Torch-based encoder. The encoder extracts audio embeddings in parallel and transfers the resulting embeddings from GPU to CPU. Subsequently, the SGLang-based LLM rollout takes control of the GPU to generate multiple hypothesis sequences based on the audio embeddings and the instruction text tokens. Each hypothesis is assigned a reward according to the predefined rules, which will be detailed later.
Finally, the FSDP-based LLM policy model uses the audio embeddings and the generated hypotheses to compute output probabilities and performs policy optimization via RL. After each update, the optimized policy is synchronized back to the rollout module, ensuring that the RL process remains on-policy.

We evaluate the training efficiency of FunRL on 8 A100 GPUs, with results shown in Figure~\ref{fig:rl_right}. For approximately one hour of input audio, one training step takes about 54.6 seconds, yielding a \textbf{real-time factor (RTF) of approximately 0.015}. As shown in Figure~\ref{fig:rl_right}, the SGLang rollout phase dominates the computation time, while device-switching overhead accounts for less than 6\% of the total computation time. This observation indicates that the strategy of alternating GPU utilization in FunRL is highly efficient, \textbf{making FunRL a scalable and effective solution for RL training in LALMs}.

% The encoder can extract in parallel。

\subsubsection{GRPO-based RL for ASR}

Based on the FunRL framework, we enhance the GRPO-based RL algorithm for Fun-ASR. 
Among various RL algorithms, GRPO
%%%%~\cite{DBLP:journals/corr/abs-2402-03300}
is a lightweight and effective policy-based method that has achieved significant success in LLM post-training. 
Unlike RL algorithms such as PPO,
%%%%~\cite{DBLP:journals/corr/SchulmanWDRK17}
GRPO generates a group of responses $\{o_i\}_{i=1}^{G}$ and computes a rule-based value function to assign a reward $\{R_i\}_{i=1}^G$ to each response. These group-level rewards are then normalized to compute the advantage $\hat{A}_{i,t}$,  which guides the policy update.

\begin{equation}
\hat{A}_{i,t}=\frac{R_i-\text{mean}\!\left(\{R_j\}_{j=1}^G\right)}{\text{std}\!\left(\{R_j\}_{j=1}^G\right)}
\end{equation}
The policy is optimized with a clipped objective and a directly imposed KL penalty term.

\begin{equation}
\begin{split}
L_{\mathrm{GRPO}}(\theta) = 
\frac{1}{G}\sum_{i=1}^G \frac{1}{|o_i|}\sum_{t=1}^{|o_i|} 
&\min\!\bigl(r_{i,t}(\theta)\,\hat{A}_{i,t},\;
\operatorname{clip}\bigl(r_{i,t}(\theta),1-\varepsilon,1+\varepsilon\bigr)\,\hat{A}_{i,t}\bigr) \\
& - \beta\,D_{\mathrm{KL}}\!\bigl(\pi_\theta\|\pi_{\mathrm{ref}}\bigr)
\end{split}
\end{equation}
where
\begin{equation}
r_{i,t}(\theta)=\frac{\pi_{\theta}\bigl(o_{i,t}\mid q,\,o_{i,<t}\bigr)}
{\pi_{\theta_{\mathrm{old}}}\bigl(o_{i,t}\mid q,\,o_{i,<t}\bigr)}.
\end{equation}

We observe that when WER is used as the value function, GRPO becomes similar to Minimum Word Error Rate (MWER), a widely adopted criterion in the ASR community. In this work, we further design a set of new value functions $\{R^k(y_i^*, y_i)\}_{k=1}^K$ to enhance both ASR performance and user experience:

\begin{itemize}[leftmargin=*,noitemsep]
    \item \textbf{ASR Accuracy ($R_i^1$)}. To directly optimize the recognition quality, we use the $1 - \text{WER}(y^*, y)$ as the basic value function, and the value range is $[0, 1]$.
    
    \item \textbf{Keyword Accuracy and Recall ($R_i^2$).} Since keywords have a significant impact on user experience, we incorporate keyword recall as a reward component. Keywords for each utterance are either manually annotated or identified by an LLM. However, using recall alone tends to increase insertion errors; therefore, we also include keyword accuracy to balance precision and recall.
    
    \item \textbf{Noise Robustness and Hallucination Suppression ($R_i^3$).} Hallucination is a common issue in LLM-based ASR systems, especially under noisy conditions. To mitigate this, we detect hallucinated content via regular expression matching and apply penalties proportional to the length of the hallucinated span.
    
    \item \textbf{Language Match ($R_i^4$).} In certain cases, the model may inadvertently produce speech translation instead of transcription. To enforce language consistency, we assign a final reward of $-1$ if the output language does not match the source language.
\end{itemize}

Except for $R_i^4$, all of the function results are summed to obtain the final $R_i$. Although $R_i^2$ to $R_i^4$ can be reflected by the ASR accuracy, our experimental results show that adding these rules significantly improves the user experience and reduces the WER on the hard cases.

% \item ASR Accuracy
% \item ASR Accuracy
% \item ASR Accuracy

% \begin{table}[]
%     \centering
%     \begin{tabular}{llc}
%     \toprule
%         $R^k(y^*, y)$ & Description & Range\\
%     \midrule
%         ASR Accuracy & $1 - \text{WER}(y^*, y)$ & $(-\inf, 1]$ \\
%         Language Match &  & $\{-\inf, 0\}$ \\
%         Keyword Recall &  & $[0, 1]$ \\
%         Keyword Accuracy &  & $[-1, 0]$ \\
%          &  & $[-1, 0]$ \\
%     \bottomrule
%     \end{tabular}
%     \caption{Caption}
%     \label{tab:placeholder}
% \end{table}

\subsubsection{Constructing the RL Training Data}

For addressing practical issues in real-world application scenarios, we construct a small but high-quality RL training data using the following approach.

\begin{itemize}[leftmargin=*,noitemsep]

\item \textbf{Hardcase Samples.} We collect a large amount of unlabeled speech and transcribe each utterance using the Fun-ASR model after contextual SFT (referred to as the base model),  along with three other distinct ASR systems, including Whisper, FireRed-ASR, and SenseVoice. If the outputs from the three external systems are consistent with each other (WER < 5\%) but differ significantly from Fun-ASR’s output (WER > 10\%), the sample is identified as a hard case and is included in the RL training set.

\item \textbf{Long-duration Samples.} We select audio segments longer than 20 seconds to improve model performance on extended speech inputs, which are common in real-world applications but scarce in the training data (less than 10\%).

\item \textbf{Hallucination-related Samples.} We specifically include data where the base model exhibits hallucination behavior, which could be significantly longer than the ground truth or contain repeated phases. In addition, we incorporate the utterances with the reference transcripts containing long repetitions of words or phrases, cases that resemble hallucinations but are genuinely present in the audio, to help the model distinguish between real and spurious patterns.

\item \textbf{Keyword and Hotword Samples.} For utterances without predefined hotwords, we use Qwen-2.5 7B to identify salient keywords. For hotword-specific training, we use the hotwords in the reference transcription as target keywords.

\item \textbf{Regular ASR Data.} A subset of standard ASR data is included to mitigate catastrophic forgetting and maintain general recognition performance during RL training.

\end{itemize}

The final RL training data consist of 100K samples, with 20K utterances in each of the five subsets described above. Thanks to the efficiency of the FunRL framework, the entire RL training of Fun-ASR can be completed within one day with 8 A100 GPUs.

 % value function and estimates the advantage in a group-relative manner.

% Among various RL algorithms, GRPO [16] provides a lightweight alternative to policy based methods such as PPO [20]. Unlike PPO [20], which requires training a separate value function—normally a separate model of similar size to the policy network, GRPO introduces a group relative formulation to estimate the advantage. In our setting, we recast the GRPO framework into the context of flow matching.

% \begin{figure}
%     \centering
%     \includegraphics[width=0.4\linewidth]{img/rl_framework.png}
%     \caption{Reinforcement learning Framework and time consumption analysis the time  of the Fun-ASR.}
%     \label{fig:rl_framework}
%     \vspace{-8mm}
% \end{figure}

% @ changfeng

\section{Production-oriented Optimization}
\label{sec:production-optimization}

\subsection{Streaming Ability}
\label{subsec:streaming}
To enhance the streaming capability of large audio language model Fun-ASR, we construct streaming-style training data that explicitly emulate the streaming decoding process, thereby reducing the mismatch between training and inference. Specifically, we sample a subset of the offline training corpus and transform it into incremental, chunked inputs that expose only past context. Fine-tuning by combining this simulated streaming with previous offline training data improves the model’s performance under streaming decoding.

\subsection{Noise Robust Training}
\label{subsec:noise-robustness}
Given the diverse range of real-world deployment scenarios, it is essential for Fun-ASR to maintain reliable performance under challenging acoustic conditions, such as those found in restaurants, train stations, and shopping malls, without significant performance degradation. However, creating a dataset that fully captures the complexity and variability of real-world noise environment is impractical. To address this challenge, we employ a large-scale noisy data augmentation strategy.
We begin by selecting approximately 110K hours of low-noise speech and 10K hours of noise samples from our in-house corpus. These are combined to generate around 110K hours of offline simulated noisy speech, with an average signal-to-noise ratio (SNR) of 10 dB and a standard deviation of 5 dB.
To further increase data diversity, 30\% of the training speech is randomly chosen for online data augmentation, where environmental noise is mixed during training. This comprehensive approach for noise robustness yields approximately \textbf{13\% average relative performance improvement} on our complex-noise evaluation set.

\subsection{Multilingual ASR}
\label{subsec:multilinguality}
The availability of training data varies widely across different languages. Resource-rich languages such as Chinese and English have abundant data, whereas languages such as Vietnamese and Thai have comparatively limited resources. The primary Fun-ASR model is a Chinese-English model. To improve multilingual ASR performance, we train an additional multilingual Fun-ASR model to support these languages. The previous multilingual Fun-ASR model (denoted by \textbf{Fun-ASR-ML}) supports Chinese, English, Vietnamese, Thai, and Indonesian. For the latest release multilingual model Fun-ASR-ML can support 31 languages, and it also support streaming speech recognition for some languages. In order to support industrial speech recognition application, we also support context bias enhanced speech recognition by using prompt. During training, we downsample the Chinese and English data used for the Chinese-English Fun-ASR model and upsample other languages data in order to balance the distribution. In total, this multilingual dataset comprises approximately 500k hours of audio. The training methodology is the same as that used for the Chinese-English Fun-ASR model (Section~\ref{sec:training}).

\subsection{Code-switching}
\label{subsec:code-switching}
Recognition of code-switched speech has always been a challenge. To optimize ASR performance on Chinese-English code-switched speech, we synthesize the code-switching training data as follows. Firstly, we collect over 40K English key words or phrases, covering common domains such as technology, education, finance, and sports. Secondly, we use the Qwen3 \citep{DBLP:journals/corr/abs-2505-09388} model to generate Chinese-English code-switched texts related to given key words randomly selected from the pool described above. Thirdly, we use a Text-to-Speech model to synthesize speech data in a variety of voices for the LLM-generated code-switched texts, and obtain the code-switched training data.

\subsection{Hotword Customization}
\label{subsec:hotword}
Within Fun-ASR, we implement a RAG-based mechanism for hotword customization. Specifically, we construct a hotword vocabulary in which each prespecified hotword is converted into a phoneme sequence (for Chinese) or a word-piece sequence (for other languages) using a predefined lexicon. During inference, we retrieve hotword candidates from the vocabulary based on the phoneme-level or the word-piece-level edit distance between the CTC hypotheses and entries in the hotword vocabulary. The retrieved hotword candidates, together with the audio input and the CTC prediction, are used as the input to the LLM, as depicted in Figure~\ref{fig:overview}, to produce the hotword-customized output.

\subsection{Hallucination Mitigation}
\label{subsec:hallucination-mitigation}
Hallucination in ASR, where an ASR system generates text that is not present in the input audio, is particularly problematic during silence, abrupt speaker interruptions, or in noisy environments where the model may produce spurious transcriptions even without speech. To mitigate hallucination, Fun-ASR adopts the following strategy. During data augmentation, we introduce zero-padding into audio signals before adding noise, thereby creating pure-noise segments. This strategy forces the model to learn recognizing noise-only inputs and aligning its outputs accordingly, hence reducing the likelihood of hallucinated text. We find that this approach helps enhance the robustness, accuracy, and stability of Fun-ASR across diverse acoustic conditions.

\section{Evaluation}
\label{sec:evaluation}

\subsection{Evaluation Setting}

We evaluate Fun-ASR and Fun-ASR-ML on both open-source ASR benchmark datasets and real-world industry evaluation sets. For the open-source evaluation, we use corresponding test sets of AIShell-1 \citep{bu2017aishell1opensourcemandarinspeech}, AIShell-2\citep{du2018aishell2transformingmandarinasr}, Librispeech\citep{7178964}, Fleurs\citep{conneau2022fleursfewshotlearningevaluation}, WeNetSpeech\citep{zhang2022wenetspeech10000hoursmultidomain}, Gigaspeech2 \citep{yang2025gigaspeech2evolvinglargescale} data sets. These open-source datasets have been publicly available for a long time, increasing the risk of data leakage into model training sets. To ensure a more reliable and leakage-free evaluation, we collect newly uploaded videos from YouTube and Bilibili posted after June 30th, which are then manually transcribed to form an independent test set.
For noise robustness evaluation, we use real-world audio recordings captured in various environments, including canteen, dinner, meeting, office, outdoor, park, shop, street, subway, supermarket, and walk-street. These are further categorized by acoustic conditions and topic to better assess performance under diverse and challenging scenarios.

\subsection{Evaluation results}

\subsubsection{Overall Results}

We first evaluate recently published ASR systems on open-source benchmarks, with results shown in Table \ref{tab:asr_os}. On these datasets, all models achieve very low WER, and some open-source models even outperform commercial APIs on Librispeech and AIShell. However, as shown in Table~\ref{tab:asr_industry}, on real industry evaluation sets, Seed-ASR-API demonstrates a clear advantage over other open-source models, particularly in noisy conditions. This indicates that performance on open-source test data may not reliably reflect real-world ASR capabilities, highlighting the importance of regularly updating evaluation sets to prevent data leakage. Compared to both open-source models and commercial APIs, our Fun-ASR achieves SOTA performance on both open-source benchmarks (Table~\ref{tab:asr_os}) and industry datasets (Table~\ref{tab:asr_industry}). Since all training data is collected before June 30th 2025, we ensure no data leakage during our evaluations, making the results trustworthy and reproducible.
Notably, Fun-ASR-nano also outperforms the open-source models and performs closely to Seed-ASR, with only \textbf{0.8B} parameters.
% \wen{Please discuss the results in more depth, i.e., if all results are positive gains, please report the relative gains. If some results are not as expected, please provide our explanations or hypothesis, and insights on how to improve our results.}
% @pengzhendong
\begin{table}[htbp]
    \centering
    \begin{adjustbox}{max width=\textwidth}
    \begin{tabular}{lcccccccccc}
    \toprule
             & GLM-ASR & GLM-ASR & Whisper- & Seed- & Seed- & Kimi- & Step- & FireRed- & Fun-ASR & Fun-\\
    Test set & -nano & -nano* & large-v3 & ASR & ASR* & Audio & Audio2 & ASR & -nano & ASR\\
     \midrule
     Model Size & 1.5B & 1.5B & 1.6B & - & - & - & - & 1.1B & 0.8B & 7.7B \\
     OpenSource & \textcolor{green}{\cmark} & \textcolor{green}{\cmark} & \textcolor{green}{\cmark} & \textcolor{red}{\xmark} & \textcolor{red}{\xmark} & \textcolor{green}{\cmark} & \textcolor{green}{\cmark} & \textcolor{green}{\cmark} & \textcolor{green}{\cmark} & \textcolor{red}{\xmark} \\     
     AIShell1 & 1.81 & 2.17 & 4.72 & 0.68 & 1.63 & 0.71 & 0.63 & 0.54  & 1.80 & 1.22\\
     AIShell2 & - & 3.47 & 4.68 & 2.27 & 2.76 & 2.86 & 2.10 & 2.58 & 2.75 & 2.39\\
     Fleurs-zh & - & 3.65 & 5.18 & 3.43 & 3.23 & 3.11 & 2.68 & 4.81 & 2.56 & 2.53\\
     Fleurs-en & 5.78 & 6.95 & 6.23 & 9.39 & 9.39 & 6.99 & 3.03 & 10.79  & 5.96 & 4.74\\
     Librispeech-clean & 2.00 & 2.17 & 1.86 & 1.58 & 2.8 & 1.32 & 1.17 & 1.84 & 1.76 & 1.51\\
     Librispeech-other & 4.19 & 4.43 & 3.43 & 2.84 & 5.69 & 2.63 & 2.42 & 4.52 & 4.33 & 3.03\\
     WenetSpeech Meeting & 6.73 & 8.21 & 18.39 & 5.69 & 7.07 & 6.24 & 4.75 & 4.95  & 6.60 & 6.17\\
     WenetSpeech Net & - & 6.33 & 11.89 & 4.66 & 4.84 & 6.45 & 4.67 & 4.94 & 6.01 & 5.46\\
    \bottomrule
    \end{tabular}
    \end{adjustbox}
    \vspace{3mm}
    \caption{Evaluation results in terms of WER (\%) on \textbf{open-source datasets}. The Seed-ASR* results are evaluated using the official API on the volcengine and the GLM-ASR-nano* results are evaluated using the opensource checkpoint.}
    % \wen{I modified the caption of this table, (1) please modify the captions of other tables to highlight the test sets and evaluation metrics. Also, (2) please boldface the best results in table for each test set or test condition. (3) Please add the model size right under each model **the first time** we compare these models in tables, so that readers can clearly compare the model sizes. (4) Please be consistent with Fun-ASR. We used Fun-ASR in many places...}}
    \label{tab:asr_os}
\end{table}

% @pengzhendong
\begin{table}[htbp]
    \centering
    \begin{adjustbox}{max width=\textwidth}
    \begin{tabular}{lccccccccc}
    \toprule
             & GLM-ASR & Whisper-v3 & Seed- & FireRed- & Kimi- & Paraformer & Fun-ASR & Fun-\\
    Test set & -Nano & -large & ASR & ASR & Audio & v2 & -nano & ASR\\
     \midrule
     Model Size & 1.5B & 1.6B & - & 1.1B & 8B & 0.2B & 0.8B & 7.7B \\
     OpenSource & \textcolor{green}{\cmark} & \textcolor{green}{\cmark} & \textcolor{red}{\xmark} & \textcolor{green}{\cmark} & \textcolor{green}{\cmark} & \textcolor{green}{\cmark} & \textcolor{green}{\cmark} & \textcolor{red}{\xmark} \\
     Nearfield & 16.95 & 16.58 & 7.20 & 10.10 & 9.02 & 8.11 & 7.79 & 6.31\\
     Farfield & 9.44 & 22.21 & 4.59 & 7.49 & 10.95 & 9.55 & 5.79 & 4.34\\
     Complex Background & 23.79 & 32.57 &  12.90 & 15.56 & 15.56 & 15.19 & 14.59 & 11.45\\
     English General & 16.47 & 18.56 & 15.65 & 21.62 & 18.12 & 19.48 & 15.28 & 13.73\\
     Opensource & 4.67 & 7.05 & 3.83 & 5.31 & 3.79 & 6.23 & 4.22 & 3.38\\
     Dialect & 54.21 & 66.14 & 29.45 & 52.82 & 71.94& 41.16 & 28.18 & 15.21\\
     Accent & 19.78 & 36.03 & 10.23 & 14.05 & 27.20 & 17.80 & 12.90 & 10.31\\
     Lyrics & 46.56 & 54.82 & 30.26 & 42.87 & 65.18 & 50.14 & 30.85 & 21.00\\
     Hiphop & 43.32 & 46.56 & 29.46 & 33.88 & 57.25 & 43.79 & 30.87 & 28.58\\
     \midrule
    Average & 26.13 & 33.39 & 15.95  & 22.63 & 31.00 & 23.49 & 16.72 & 12.70 \\

    \bottomrule
    \end{tabular}
    \end{adjustbox}
    \vspace{3mm}
    \caption{Evaluation results in terms of WER (\%) on \textbf{Industry Datasets}.}
    \label{tab:asr_industry}
\end{table}

\subsubsection{Streaming ASR Performance}

In order to evaluate the streaming ability of our Fun-ASR model, we evaluate the performance on the same test set used when evaluating the offline speech recognition ability of our Fun-ASR model. Table \ref{tab:streaming_asr_performance} lists the testing results. When comparing with Seed-ASR ~\citep{DBLP:journals/corr/abs-2407-04675}, our Fun-ASR model have better performance at different test sets and test scenario.

\begin{table}[htbp]
    \centering
    \begin{adjustbox}{max width=\textwidth}
    \begin{tabular}{lccccccccc}
    \toprule
    Test set & Seed-ASR & Fun-ASR-nano & Fun-ASR\\
     \midrule
     Nearfield & 8.64 & 8.10 & 6.75 \\
     Farfield & 5.51 & 6.38 & 4.72\\
     % Home Scenario & 9.7 & 6.84 & 5.33\\
     % Home Scenario &  \\
      % & 10.48 & 10.00 & 12.34 & 11.09 \\
     Complex Background  & 15.48 & 15.52 & 12.49\\
     English General & 18.78 & 16.46 & 14.68\\
     OpenSource & 3.80 & 5.06 & 4.08\\
     Dialect & - & 30.72 & 18.25\\
     Accent & - & 15.42 &  11.49\\
     Lyrics & - & 31.54 &  22.05\\
     Hiphop & - & 36.55 &  28.90\\
     % \midrule
    % Average & 10.32 & 9.73 & 8.08\\
    \bottomrule
    \end{tabular}
    \end{adjustbox}
    \vspace{3mm}
    \caption{Evaluation results in terms of WER (\%) with \textbf{streaming decoding}.}
    \label{tab:streaming_asr_performance}
\end{table}

% @nichongjia
% \begin{table}[htbp]
%     \centering
%     \begin{adjustbox}{max width=\textwidth}
%     \begin{tabular}{lcccccccc}
%     \toprule
%     Test set & Qwen-ASR & Seed-ASR & Kimi-Audio & Step-Audio2 & FireRed-ASR & Fun-ASR & Fun-ASR-mini \\
%      \midrule

%     \bottomrule
%     \end{tabular}
%     \end{adjustbox}
%     \vspace{3mm}
%     \caption{Word Error Rate (WER, \%) evaluation Result on Hotword.}
%     \label{tab:multilingual_asr_performance}
% \end{table}

\subsubsection{Evaluation on Noise Robustness}

\begin{table}[htbp]
    \centering
    \begin{adjustbox}{max width=\textwidth}
    \begin{tabular}{lcccc}
    \toprule
    % Test set & \multicolumn{3}{c}{Name} & \multicolumn{3}{c}{Since} & \multicolumn{3}{c}{Name} \\
    %  &  WER & rec & acc & WER & rec & acc & WER & rec & acc \\
     & \multicolumn{3}{c}{Fun-ASR} & \\
    Environment & w/o NRT & w/ NRT & NRT + RL\\
     \midrule
     canteen & 20.67 & 19.66 & 19.28\\
     dinner & 14.02 & 9.02 & 8.77\\
     meeting & 6.45 & 6.25 & 6.05\\
     office & 15.02 & 10.52 & 10.42\\
     outdoor & 10.12 & 9.63 & 9.37\\
     park & 13.67 & 11.04 & 10.89\\
     shop & 12.22 & 10.57 & 10.46\\
     street & 12.05 & 10.1 & 9.85\\
     subway & 14.11 & 11.89 & 11.84\\
     supermarket & 14.27 & 8.03 & 7.74\\    
     walkstreet & 13.89 & 13.34 & 12.94\\
     % &  & 6.55 &  & 7.24 \\
     % &  & 11.58 &  & 13.07 \\
     % &  & 5.14 &  & 6.96 \\
     % &  & 5.19 &  & 6.53  \\
     % &  & 10.48 &  & 12.34  \\
     % &  & 12.16 &  & 13.53 \\
     % &  & 15.17 &  & 15.54 \\
     % &  & 3.98 &  & 5.17  \\
     
     \midrule
    Average & 13.32 & 10.91 & 10.69\\
    \bottomrule
    \end{tabular}
    \end{adjustbox}
    \vspace{3mm}
    \caption{\textbf{Noise robust evaluation} under different environments.}
    \label{tab:asr_noise}
\end{table}

We present the noise robustness evaluation in Table \ref{tab:asr_noise}. It is evident that noise robust training (NRT) is crucial for industrial applications. In challenging environments such as dinner and supermarket settings, NRT brings over 30\% relative improvement, as LLM-based ASR systems tend to generate hallucinated outputs under such complex acoustic conditions. Furthermore, RL further enhances the model's noise robustness.

\subsubsection{Code-switching Evaluation}

For evaluation, two test sets A and B are used to evaluate the effectiveness of the constructed code-switched training data (Section~\ref{subsec:code-switching}). Test sets A and B are randomly selected from the 
daily\_dialogue\_mixed\_chinese\_english\_speech\_tts dataset \citep{zhendataset} and our in-house recordings, respectively. The results are shown in Table~\ref{tab:asr_cs}.

\begin{table}[htbp]
    \centering
    \begin{adjustbox}{max width=\textwidth}
    \begin{tabular}{ccccccc}
    \toprule
    % Test set & \multicolumn{3}{c}{Name} & \multicolumn{3}{c}{Since} & \multicolumn{3}{c}{Name} \\
    %  &  WER & rec & acc & WER & rec & acc & WER & rec & acc \\
    Test set & \multicolumn{3}{c}{Offline} & \multicolumn{3}{c}{Streaming}  \\
     & w/o CS & w/o RL & w/ RL & w/o CS & w/o RL & w/ RL   \\
     \midrule
     A & 4.53 & 1.70 & 1.55 & 6.19 & 5.85 & 2.28\\
     \midrule
     B & 4.76 & 4.50 & 4.49 & 6.32 & 5.68 & 5.07\\
    \bottomrule
    \end{tabular}
    \end{adjustbox}
    \vspace{3mm}
    \caption{Evaluation results in terms of WER (\%) on \textbf{code-switched test sets}.}
    \label{tab:asr_cs}
\end{table}

\subsubsection{Evaluation on Hotword Customization}

% @pengzhendong
\begin{table}[htbp]
    \centering
    \begin{adjustbox}{max width=\textwidth}
    \begin{tabular}{lcccccccccccc}
    \toprule
    Topic & \multicolumn{3}{c}{Offline w/o RL} & \multicolumn{3}{c}{Offline w/RL} & \multicolumn{3}{c}{Streaming w/o RL} & \multicolumn{3}{c}{Streaming w/ RL} \\
     &  WER & acc & rec & WER & acc & rec & WER & acc & rec & WER & acc & rec \\
     \midrule
    biology &  1.67 & 0.98 & 0.99 & 1.70 & 0.97 & 1.00 & 2.04& 0.98& 0.98 & 1.97& 0.99& 0.98 \\
    math & 0.86& 0.99& 0.99 & 0.86& 0.99& 0.99 & 1.29& 0.99& 0.99 & 1.01& 0.99& 1.00 \\
    religion & 3.20& 0.98& 0.98 & 2.87& 0.99& 0.99 & 3.71& 0.99& 0.98 & 3.35& 0.99& 0.97 \\
    food & 1.90 & 0.98& 0.99 & 1.55& 0.99& 1.00 & 2.01& 0.99& 0.99 & 1.47& 0.99& 0.99 \\
    name & 0.53& 1.00& 0.95 & 0.35& 1.00& 1.00 & 1.29& 1.00& 0.95 & 0.88& 1.00& 0.98 \\
    brand &  0.41& 1.00& 0.99 & 0.33& 1.00& 0.99 & 1.08& 0.99& 0.95 & 0.38& 1.00& 1.00 \\
    astronomy & 2.11& 1.00& 0.97 & 1.97& 0.99& 0.97 & 2.28& 0.98& 0.95 & 2.39& 1.00& 0.98 \\
    chemistry & 1.76& 0.99& 0.97 & 1.91& 0.99& 0.98 & 2.81& 0.98& 0.97 & 1.83& 0.99& 0.97 \\
    philosophy & 3.03& 0.99& 0.96 & 2.84& 0.99& 0.97 & 3.31& 0.99& 0.96 & 3.03& 0.99& 0.95 \\
    physics & 1.72& 0.99& 1.00 & 1.82& 0.98& 1.00 & 2.31& 0.99& 0.98 & 1.8& 0.99& 0.99 \\
    \bottomrule
    \end{tabular}
    \end{adjustbox}
    \vspace{3mm}
    \caption{\textbf{Hotword customization} comparison between the models w/ or w/o reinforcement learning.}
    \label{tab:rl_hotword}
\end{table}

For the hotword evaluation, we choose the audios with some special topics, including biology, math, religion, food, name, astronomy, chemistry, philosophy, and physics, as the recognition of the technical terms is crucial but still challenging for most ASR systems. Results in Table \ref{tab:rl_hotword} shows that Fun-ASR can benefit from the hotword customization. On most topics, the recall rate can raise to more than 0.97 for Fun-ASR. The Fun-ASR shows good performance on the name topic, the recall can increase from 0.75 to 1.0. This indicates the hotword customization can really inspire the target keyword, rather than just provide contextual information.

\subsubsection{Multilingual ASR Results}
We also evaluate our multilingual ASR model Fun-ASR-ML on several open source test sets and in-house industry test sets. Table \ref{tab:multilingual_asr_performance} lists the testing results. From Table \ref{tab:multilingual_asr_performance}, we can see that, on the Chinese and English open source test sets and in-house industry test sets, our multi-lingual ASR model Fun-ASR-ML have better or comparable effects when comparing with Kimi-Audio ~\citep{kimiteam2025kimiaudiotechnicalreport}. We also compared our model with other multi-lingual ASR models, such as Whisper large v3~\citep{DBLP:conf/icml/RadfordKXBMS23}, dolphin-small ~\citep{meng2025dolphinlargescaleautomaticspeech}, and seamless-m4t large v2 ~\citep{communication2023seamlessmultilingualexpressivestreaming}. When comparing with these models, our Fun-ASR-ML model can also get the SOTA performance. In Table \ref{tab:multilingual_asr_performance}, we also evaluate our multilingual ASR Fun-ASR-ML-Nano model on several open source test sets and in-house industry test sets. Although when comparing our Fun-ASR-ML model, the performance of Fun-ASR-ML-Nano degraded a little bit, it also get better performance when comparing with other models.

\begin{table}[htbp]
    \centering
    \begin{adjustbox}{max width=\textwidth}
    \begin{tabular}{lcccccccc}
    \toprule
    Language & Test set & Kimi-Audio & Whisper Large v3 & dolphin-small & seamless-m4t-large-v2 & Fun-ASR-ML & Fun-ASR-ML-Nano \\
     \midrule
     \multirow{5}{*}  & fleurs & 2.69 &  4.71 & 5.46 & 5.15 & 3.0 & 3.51\\
                            & commonvoice & 7.21 & 12.61 & 9.94 & 10.76 & 5.76 & 6.2\\
                 {Chinese}  & wenetspeech-test-net & 5.37 & 9.83 & 9.63 & 9.87 & 6.48 & 6.35\\
                            & aishell2-ios-test & 2.56 & 4.83& 4.37 & 4.79 &2.60 & 2.74\\
                            & in-house-test-set & 36.42 & 16.54 & 9.67 & 14.85 & 7.91 & 7.89\\
    \midrule
    \multirow{5}{*}            & fleus & 4.4 & 4.11 & N.A. & 6.59 & 3.18 & 5.49 \\
                            & commonvoice & 10.31 & 9.66 & N.A. & 7.63 & 7.67 & 9.90 \\
                 {English}  & librispeech-test-clean & 1.28 & 2.56 & N.A. & 2.56 & 1.62 & 1.68 \\
                            & librispeech-test-other & 2.42 & 4.34 & N.A. & 4.84 & 3.39 & 4.03 \\
                            & in-house-test-set & 12.40 & 11.78 & N.A.& 43.74 & 11.19 & 11.61 \\
    \midrule
    \multirow{4}{*}              & fleurs & N.A. & 6.07 & 15.86 & 9.36 & 8.10 & 6.56\\
                {Indonesian}  & commonvoice & N.A. &7.27 & 8.91 & 6.1 & 5.49 & 8.09 \\
                              & gigaspeech2-test & N.A. &19.11 & 26.56 & 22.3 & 16.71 & 18.52\\
                              & in-house-test-set & N.A. & 23.19 & 40.16 & 24.41 & 18.36 & 21.42\\
    \midrule
    \multirow{4}{*}         & fleurs & N.A. & 8.48 & 9.66 & 9.25 & 6.0 & 8.05 \\
                {Thai}   & commonvoice & N.A. & 5.92 & 3.04 & 2.81 & 0.9 & 1.97\\
                         & gigaspeech2-test & N.A. & 19.35 & 19.15 & 21.7 & 19.09 & 18.96 \\
    \midrule
    \multirow{4}{*}              & fleurs & N.A. & 6.51 & 15.62 & 8.07 & 5.50 & 7.75\\
                {Vietnamese}  & commonvoice & N.A. & 13.51 & 12.73 & 13.85 & 7.41 & 9.26\\
                              & gigaspeech2-test & N.A. & 13.82 & 31.98 & 43.31 & 8.98 &9.29 \\
                              & in-house-test-set & N.A. & 11.46 & 40.82 & 32.10 & 7.06 & 7.72\\
    \bottomrule
    \end{tabular}
    \end{adjustbox}
    \vspace{3mm}
    \caption{Evaluation results in terms of WER (\%) or CER (\%) on different \textbf{multilingual test sets}.}
    \label{tab:multilingual_asr_performance}
\end{table}

\subsubsection{Effect of Reinforcement Learning}

Table \ref{tab:asr_rl} shows that RL plays a crucial role in Fun-ASR training, bringing approximately 4.1\% and 9.2\% relative improvement under offline and streaming conditions, respectively. For offline ASR, the performance gain is more pronounced on audio from noisy and complex environments compared to clean or open-source data. Notably, the improvement is even more significant in the streaming ASR setting. RL helps suppress both insertion and deletion errors, which may be attributed to early termination or premature prediction before full pronunciation.

\begin{table}[htbp]
    \centering
    \begin{adjustbox}{max width=\textwidth}
    \begin{tabular}{lcccccc}
    \toprule
    % Test set & \multicolumn{3}{c}{Name} & \multicolumn{3}{c}{Since} & \multicolumn{3}{c}{Name} \\
    %  &  WER & rec & acc & WER & rec & acc & WER & rec & acc \\
    Test set & \multicolumn{2}{c}{Offline} & \multicolumn{2}{c}{Streaming}  \\
     & w/o RL & w/ RL & w/o RL & w/ RL \\
     \midrule
     Nearfield & 6.40 & 6.31 & 6.90 & 6.75\\
     % Outhouse & 11.58 & 11.45 & 13.07 & 12.28 \\
     Farfield & 4.42 & 4.34 & 5.03 & 4.72\\
     % Home Scenario & 5.19 & 5.17 & 5.18 & 6.53 & 5.73 & 6.72\\
      % & 10.48 & 10.00 & 12.34 & 11.09 \\
     Complex Background  & 11.71 & 11.45 & 12.93 & 12.49\\
     English Genearl & 14.11 & 13.73 & 15.17 & 14.68\\
     Opensource & 3.72 & 3.38 & 4.19 & 4.08\\
     Dialect & 15.51 & 15.21 & 18.64 & 18.25\\
     Accent & 10.60 & 10.31 & 11.78 & 11.49\\
     Lyrics & 22.23 & 21.00 & 22.51 & 22.05\\
     Hiphop & 29.96 & 28.58 & 29.92 &  28.90\\
     \midrule
    Average & 13.18 & 12.70 & 14.12 
 & 13.71 \\
    \bottomrule
    \end{tabular}
    \end{adjustbox}
    \vspace{3mm}
    \caption{Comparison between the models w/ or w/o reinforcement learning.}
    \label{tab:asr_rl}
\end{table}

As shown in Table \ref{tab:rl_hotword}, RL effectively enhances hotword integration, leading to improvements in both accuracy and recall across most test sets. In certain domains, such as philosophy and religion, the RL model may achieve slightly lower accuracy or recall compared to the baseline; however, the overall WER still decreases. This is because, during RL training, keywords are selected based on the actual transcriptions rather than the input prompts, enabling Fun-ASR to better recognize domain-specific terminology—even for professional terms not explicitly included in the hotword list.

\section{Limitations and Future Plans}

Despite strong results across diverse evaluations, our Fun-ASR model still has some limitations. First, it is primarily optimized for Chinese and English—particularly for streaming performance and hotword customization—so support for other languages remains limited. Second, the effective context window is constrained; without an external voice activity detection (VAD) module, the system struggles to handle long-duration recordings robustly. Third, the current release does not support far-field or multi-channel audio. We plan to address these limitations in future work.

\section{Conclusion}

In this paper, we present Fun-ASR, a large-scale, LLM-based automatic speech recognition (ASR) system that leverages massive data, extensive model capacity, seamless integration of LLMs, and refinement learning to achieve state-of-the-art performance across diverse and challenging scenarios. Designed with practical deployment in mind, Fun-ASR incorporates key optimizations for real-world applications, including enhanced streaming capabilities, robustness to noise, effective handling of code-switching, and customizable hotword support. Experimental results reveal that while many LLM-based ASR systems perform well on open-source benchmarks, they often underperform on real-world industrial evaluation sets. In contrast, Fun-ASR—through production-oriented design and optimization—demonstrates superior accuracy on practical application datasets, establishing a new benchmark for high-performance, deployable ASR systems.

\section{Authors (in alphabetical order of last name)}
\begin{multicols}{3}
	\begin{itemize}[noitemsep]
		\item Keyu An
            \item Yanni Chen
            \item Zhigao Chen
            \item Chong Deng
            \item Zhihao Du
		\item Changfeng Gao
		\item Zhifu Gao
            \item Bo Gong
            \item Xiangang Li
            \item Yabin Li
            \item Ying Liu
		\item Xiang Lv
            \item Yunjie Ji
            \item Yiheng Jiang
            \item Bin Ma
            \item Haoneng Luo
            \item Chongjia Ni
            \item Zexu Pan
            \item Yiping Peng
            \item Zhendong Peng
            \item Peiyao Wang
            \item Hao Wang
            \item Haoxu Wang
		      \item Wen Wang
            \item Wupeng Wang
            \item Yuzhong Wu
            \item Biao Tian
            \item Zhentao Tan
            \item Nan Yang
            \item Bin Yuan
            \item Jieping Ye
            \item Jixing Yu
            \item Qinglin Zhang
            \item Kun Zou
            \item Han Zhao
            \item Shengkui Zhao
            \item Jingren Zhou   
            \item Yanqiao Zhu
	\end{itemize}
\end{multicols}

\section{Acknowledgment}

We are immensely grateful for the invaluable discussions, support, and assistance we received from many colleagues during the development. Special thanks go to:
Mengzhe Chen,
Yafeng Chen,
Yuezhang Wang.

\bibliographystyle{iclr2023_conference}
\bibliography{refs}
\end{document}